\documentclass[preprint,12pt]{elsarticle}

\usepackage{amsmath,amssymb,amsfonts,amsthm}
\usepackage{algorithm, algorithmic}
\usepackage{graphicx}
\usepackage{textcomp}
\usepackage{xcolor}
\usepackage{dsfont}
\usepackage{comment}
\usepackage{hyperref}
\usepackage{tikz}
\usepackage{xspace}
\usepackage{chngcntr}
\usepackage{apptools}
\usepackage[symbol]{footmisc}

\usetikzlibrary{positioning}
\tikzstyle{obs} = [circle, draw=black, fill=yellow!30, thick, minimum size=7mm]
\tikzstyle{state} = [circle, draw=black, fill=green!30, thick, minimum size=7mm]
\tikzstyle{switch} = [rectangle, draw=black, fill=blue!30, thick, minimum size=7mm]
\tikzstyle{cache} = [rectangle, draw=black, fill=purple!30, thick, minimum size=7mm]
\tikzstyle{label} = [rectangle, draw=black, fill=white, thick, minimum size=7mm]
\tikzstyle{input} = [circle, draw=black, fill=blue!30, thick, minimum size=12mm]

\newcommand{\bra}[1]{\left({#1}\right)}
\newcommand{\sbra}[1]{\left[{#1}\right]}
\newcommand{\cbra}[1]{\left\{{#1}\right\}}
\newcommand{\abs}[1]{\left|{#1}\right|}

\newcommand{\eE}[1]{\mathbb{E}\sbra{{#1}}}
\newcommand{\var}[1]{\text{Var}\sbra{#1}}
\newcommand{\norm}[1]{\left\lVert{#1}\right\rVert}

\newcommand\defines{\mathrel{\stackrel{\makebox[0pt]{\mbox{\normalfont\tiny def}}}{=}}}
\newcommand{\grad}{\nabla_{\theta}}

\makeatletter
\DeclareRobustCommand\onedot{\futurelet\@let@token\@onedot}
\def\@onedot{\ifx\@let@token.\else.\null\fi\xspace}

\def\ie{\emph{i.e}\onedot} \def\Ie{\emph{I.e}\onedot}

\makeatother

\AtAppendix{\counterwithin{corollary}{section}} 
\newtheorem{corollary}{Corollary}
\AtAppendix{\counterwithin{lemma}{section}}

\AtAppendix{\counterwithin{theorem}{section}}
\newtheorem{theorem}{Theorem}

\journal{Signal Processing}

\begin{document}

\begin{frontmatter}



\title{Differentiable Interacting Multiple Model Particle Filtering}


\author[CSSurrey,NPL]{John-Joseph Brady\footnote[1]{Corresponding author, john-joseph.brady@kcl.ac.uk. J. Brady has transferred institutions to King's College London since this work was submitted for review.}}

\author[NPL]{Yuhui Luo}
\author[CVSSP]{Wenwu Wang}
\author[Ed]{V\'{i}ctor Elvira}
\author[CVSSP,COCTS]{Yunpeng Li}

\affiliation[CSSurrey]{organization={Computer Science Research Centre, University of Surrey},
            city={Guildford},
            state={Surrey},
            country={United Kingdom}}
\affiliation[NPL]{organization={Data Science Department, National Physical Laboratory},
            city={Teddington},
            state={Greater London},
            country={United Kingdom}}
\affiliation[CVSSP]{organization={Centre for Vision, Speech and Signal Processing, University of Surrey},
            city={Guildford},
            state={Surrey},
            country={United Kingdom}}
\affiliation[Ed]{organization={School of Mathematics, University of Edinburgh},
            city={Edinburgh},
            country={United Kingdom}}
\affiliation[COCTS]{organization={Centre for Oral, Clinical \& Translational Sciences, King's College London},
            city={London},
            country={United Kingdom}}
\begin{abstract}
We propose a sequential Monte Carlo algorithm for parameter learning when the studied model exhibits random discontinuous jumps in behaviour. To facilitate the learning of high dimensional parameter sets, such as those associated to neural networks, we adopt the emerging framework of differentiable particle filtering, wherein parameters are trained by gradient descent. We design a new differentiable interacting multiple model particle filter to be capable of learning the individual behavioural regimes and the model which controls the jumping simultaneously. In contrast to previous approaches, our algorithm allows control of the computational effort assigned per regime whilst using the probability of being in a given regime to guide sampling. Furthermore, we develop a new gradient estimator that has a lower variance than established approaches and remains fast to compute, for which we prove consistency. We establish new theoretical results of the presented algorithms and demonstrate superior numerical performance compared to the previous state-of-the-art algorithms.
\end{abstract}


\end{frontmatter}


\section{Introduction}
There has been longstanding interest in Bayesian filtering for systems exhibiting discontinuous behavioural jumps, typically modelled by ascribing the system a finite number of distinct and indexed regimes. Two systems frequently modelled in this way include financial markets reacting swiftly to economic news \cite{guidolin2011markovswitching, hamilton1989MS}, and tracked targets suddenly changing course or acceleration \cite{mcginnity2000MMBF, Blom1985IMM, Blom2007exactIMM, Boers2003IMM}. Much of this existing work is focused on Markov switching systems where the probability of jumping is allowed to depend only on the index of the current regime. 

Particle filters \cite{Gordon1993bootstrap, Doucet2009tutorial} are a class of Monte-Carlo algorithms for estimating the posterior distribution of a Markov hidden signal given noisy observations of it. In the regime-switching setting, if the regime index is modelled as a Markov chain, one may treat it as a component of the hidden signal in a particle filter \cite{mcginnity2000MMBF}.

In \cite{El-Laham2021RSPF} the authors developed the regime switching particle filter (RSPF), extending the approach in \cite{mcginnity2000MMBF} to systems where the regime index can depend arbitrarily on its past. They achieved this by having every particle keep a memory of its entire regime history, similar to the fixed-lag smoother of \cite{Doucet2004fixedlag}. The interacting multiple model particle filter (IMMPF), introduced in \cite{Blom2007exactIMM}, assumes the regime index is a Markov chain, but allows it to depend on the latent state as well as the index at the previous time-step. We show that, under the reformulation of the non-Markov switching model that we develop in this paper, the former problem is a special case of the latter.

In many real world settings, the average number of time-steps a system spends in each regime can be large. This is typically modelled by taking switches to be rare events. Most particle filtering algorithms naturally focus computation on more likely regions of the state space. With a restricted particle count, overtime this can result in the number of particles in all regimes apart from the current one going to zero; so when jumps do occur they are not detected \cite{Boers2003IMM, gordon2002model_uncertainty}. It has become common practice, therefore, in regime switching filters to set the number of particles assigned per regime at each time-step to be equal on average. This is achieved in \cite{El-Laham2021RSPF} by proposing the regime index uniformly across all regime choices. However, this ignores the probability of each particle adopting the given regime.

 The IMMPF takes a more principled approach, it combines the resampling and regime selection steps to improve sampling efficiency. However, the IMMPF is not strictly a particle filter in the sense studied in \cite{crisan2002survey, chopin2020book, delmoral2004book}. To the best of our knowledge, no proof of consistency for the IMMPF exists in the literature.

Differentiable particle filters (DPFs) \cite{jonschkowski2018DPF, karkus2018DPF, chen2023overview} are an emerging class of particle filters designed in such a way that the algorithm is end-to-end differentiable, so that one may obtain accurate gradient estimates for use in gradient based parameter inference. The motivating use case for DPFs is to learn model components as flexible neural networks, typically when the prior knowledge on the functional form of the underlying model is of poor quality. In this case, other parameter inference paradigms fail. For example, the EM algorithm \cite{kantas2009estimation} requires a specific functional form of the model for the maximisation step to be closed-form; and both derivative-free optimisation and particle Markov chain Monte Carlo \cite{andrieu2010PMCMC} do not scale well to large dimensional parameter spaces.

The first effort to address switching models in a DPF framework is \cite{Li2023RSDBPF}, with the regime switching differentiable bootstrap particle filter (RSDBPF). However, the RSDBPF is only capable of learning the individual regimes. The meta-model that controls the switching, henceforth the `switching dynamic', is required to be known a priori. During inference, the RSDBPF runs a RSPF so does not sample particles as efficiently as the IMMPF. Furthermore, it has an asymptotically biased gradient update.

There are few approaches in the literature that operate under an unknown switching dynamic and, to the best of our knowledge, none in the more challenging parameter estimation framework. In \cite{urteaga2016SMC_under_model_uncertainty}, a related problem is studied: the system may belong to one of a set of candidate regimes but the regime may not change during a trajectory. Their strategy is to run a separate filter for each regime but assign computational effort, \ie the number of particles, in proportion to the posterior probability that the system is in each regime. This strategy was generalised in \cite{Martino2017MMPF}, where the particles are permitted to occasionally exchange between regimes. However, this algorithm cannot provide a consistent estimator in the general case where the regime can switch at any time-step.

In this paper we propose the differentiable multiple model particle filter (DIMMPF), the first DPF approach to filtering regime-switching models where neither the individual models nor the switching dynamic are known. The DIMMPF can be seen as an IMMPF that can return statistically consistent estimates of the gradient of its filtering mean with respect to the model parameters.

The main contributions of this work\footnote{A limited version of this work was presented by the authors in the conference paper \cite{brady2024RLPF}, which presents a simpler version of our methodology that has a biased gradient update. The conference paper contains limited discussion, no theoretical insight, and a more basic set of experiments.} can be summarised as follows: 
\begin{itemize}
    \item We present the DIMMPF, a novel algorithm for learning to estimate the filtering mean of a general regime-switching model.
    \item We develop a neural network architecture to parameterise a general unknown switching dynamic.
    \item We prove that the DIMMPF generates consistent estimators of filtering means and their gradients. Entailing a derivation of, to the best of our knowledge, the first proof that filtering estimates of the IMMPF are consistent.
    \item We evaluate the DIMMPF on a set of simulated data experiments and demonstrate state-of-the-art performance.
\end{itemize}

The remainder of this article is structured as follows. In Section \ref{sec:statement} we introduce the problem statement. Section \ref{sec:background} reviews the relevant background for the paper and explains how this paper builds on previous work. Section \ref{sec:model} develops our algorithmic contribution, the DIMMPF. Section \ref{sec:experiments} describes the experiments and presents the results. We conclude in Section \ref{sec:conclusions}.

\section{Problem Statement}
\label{sec:statement}
We define a state-space model (SSM) to describe a discrete time system of two parallel processes: a latent Markov process, $\cbra{ {\hat{x}}_{t}}$; and their associated observations $\cbra{ {\hat{y}}_{t}}$, where $t$ is the discrete time index. Every observation $ {\hat{y}}_{t}$ is conditionally independent of all other variables at previous time steps given $ {\hat{x}}_{t}$. Algebraically, an SSM is defined as:
\small
\begin{equation}
\begin{gathered}
     {\hat{x}}_{0} \sim \hat{M}_{0}\bra{ {\hat{x}}_{0}}\,, \\
     {\hat{x}}_{t\geq1} \sim \hat{M}\bra{ {\hat{x}}_{t}\mid {\hat{x}}_{t-1}}\,, \\
     {\hat{y}}_{t} \sim \hat{G}\bra{ {\hat{y}}_{t}\mid {\hat{x}}_{t}}\, ,
\end{gathered}
\label{eq:SSM}
\end{equation}
\normalsize
for the set of states ${\hat{x}}_{t} \in \mathcal{X}$, the set of observations ${\hat{y}}_{t} \in \mathcal{Y}$, the random measure $\hat{M}_{0}$, and the probability kernels $\hat{M}$ and $\hat{G}$.

\begin{figure}
\centering
\begin{tikzpicture}
    \node(y0) [obs] {$ {y}_{0}$};
    \node(y1) [obs, right= 0.5cm of y0] {$ {y}_{1}$};
    \node(y2) [obs, right= 0.5cm of y1] {$ {y}_{2}$};
    \node(y3) [obs, right= 0.5cm of y2] {$ {y}_{3}$};
    \node(x0) [state, below= 0.5cm of y0] {$ {x}_{0}$};
    \node(x1) [state, below= 0.5cm of y1] {$ {x}_{1}$};
    \node(x2) [state, below= 0.5cm of y2] {$ {x}_{2}$};
    \node(x3) [state, below= 0.5cm of y3] {$ {x}_{3}$};
    \node(xdots) [state, right= 0.5cm of x3] {$\cdots$};
    \node(xT) [state, right= 0.5cm of xdots] {$ {x}_{T}$};
    \node(yT) [obs, above= 0.5cm of xT] {$ {y}_{T}$};
    \node(k0) [switch, below= 0.5cm of x0] {$k_{0}$};
    \node(k1) [switch, below= 0.5cm of x1] {$k_{1}$};
    \node(k2) [switch, below= 0.5cm of x2] {$k_{2}$};
    \node(k3) [switch, below= 0.5cm of x3] {$k_{3}$};
    \node(dots) [draw=black, fill=gray!40, thick, minimum size=0.7cm, below= 1cm of xdots] {$\cdots$};
    \node(kT) [switch, below= 0.47cm of xT] {$k_{T}$};
    \node(r0) [cache, below= 0.45cm of k0] {$ {r}_{0}$};
    \node(r1) [cache, below= 0.45cm of k1] {$ {r}_{1}$};
    \node(r2) [cache, below= 0.45cm of k2] {$ {r}_{2}$};
    \node(r3) [cache, below= 0.45cm of k3] {$ {r}_{3}$};
    \node(rT) [cache, below= 0.45cm of kT] {$ {r}_{T}$};
    \node(Label_y) [label, right = 0.5cm of yT] {Observations};
    \node(Label_x) [label, right = 0.5cm of xT] {Latent State};
    \node(Label_k) [label, right = 0.55cm of kT] {Regime Index};
    \node(Label_r) [label, right = 0.55cm of rT] {Regime Cache};

    \draw[->, thick] (x0.north) -- (y0.south);
    \draw[->, thick] (x1.north) -- (y1.south);
    \draw[->, thick] (x2.north) -- (y2.south);
    \draw[->, thick] (x3.north) -- (y3.south);
    \draw[->, thick] (xT.north) -- (yT.south);
    \draw[->, thick] (k0.north) -- (x0.south);
    \draw[->, thick] (k1.north) -- (x1.south);
    \draw[->, thick] (k2.north) -- (x2.south);
    \draw[->, thick] (k3.north) -- (x3.south);
    \draw[->, thick] (kT.north) -- (xT.south);
    \draw[->, thick] (x0.east) -- (x1.west);
    \draw[->, thick] (x1.east) -- (x2.west);
    \draw[->, thick] (x2.east) -- (x3.west);
    \draw[->, thick] (x3.east) -- (xdots.west);
    \draw[->, thick] (xdots.east) -- (xT.west);
    \draw[->, thick] (k3.east) to[out = 0, in = 180] (dots.west);
    \draw[->, thick] (dots.east) to[out = 0, in = 180] (kT.west);
    \draw[->, thick] (r0.east) -- (r1.west);
    \draw[->, thick] (r1.east) -- (r2.west);
    \draw[->, thick] (r2.east) -- (r3.west);
    \draw[->, thick] (r3.east) to[out = 0, in = 180] (dots.west);
    \draw[->, thick] (dots.east) to[out = 0, in = 180] (rT.west);
    \draw[->, thick] (k0.south) -- (r0.north);
    \draw[->, thick] (k1.south) -- (r1.north);
    \draw[->, thick] (k2.south) -- (r2.north);
    \draw[->, thick] (k3.south) -- (r3.north);
    \draw[->, thick] (kT.south) -- (rT.north);
    \draw[->, thick] (r0.east) to[out = 0, in = 180] (k1.west);
    \draw[->, thick] (r1.east) to[out = 0, in = 180] (k2.west);
    \draw[->, thick] (r2.east) to[out = 0, in = 180] (k3.west);
    \draw[->, thick] (r0.east) to[out = 0, in = 180] (k1.west);
    \draw[->, thick] (k0.north west) to[out = 180, in = 180, distance = 0.5cm] (y0.west);
    \draw[->, thick] (k1.north west) to[out = 180, in = 180, distance = 0.5cm] (y1.west);
    \draw[->, thick] (k2.north west) to[out = 180, in = 180, distance = 0.5cm] (y2.west);
    \draw[->, thick] (k3.north west) to[out = 180, in = 180, distance = 0.5cm] (y3.west);
    \draw[->, thick] (kT.north west) to[out = 180, in = 180, distance = 0.5cm] (yT.west);

\end{tikzpicture}

        \caption{Bayesian network representation of the considered regime switching model.}
        \label{fig:model}
\end{figure}
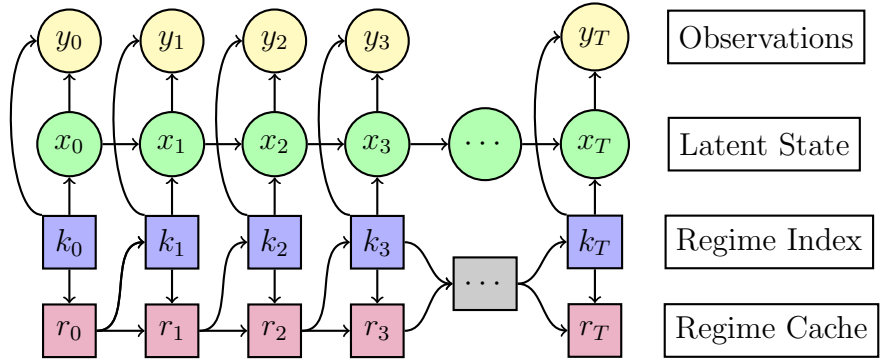

 We consider an SSM where at each time-step the latent and observation processes may, together, adopt one of a set of $N_{\text{reg}}$ regimes. To model this system, we introduce two additional latent variables: the regime index, $k_{t} \in \mathcal{K} := \cbra{1, 2, \dots, N_{\text{reg}}}$; and a cache that acts as a memory of previous regimes, $ {r}_{t} \in \mathcal{R} \subseteq \mathbb{R}^{d_{r}}$, where $d_r$ is the chosen dimension of the regime cache. We illustrate this system graphically in Fig. \ref{fig:model} and define it algebraically as:
\small
\begin{equation}
\begin{gathered}
         {r}_{0} = R^{\theta}_{0}\bra{k_{0}}\, , \\
         {r}_{t\geq1} = R^{\theta}\bra{k_{t},  {r}_{t-1}}\, , \\
        k_{0} \sim K^{\theta}_{0}\bra{k_{0}}\, ,\\
        k_{t\geq 1} \sim K^{\theta}\bra{k_{t}\mid {r}_{t-1}}\, ,\\
         {x}_{0} \sim M^{\theta}_{0}\bra{ {x}_0\mid k_{0}}\, ,\\
         {x}_{t\geq 1} \sim M^{\theta}\bra{ {x}_{t}\mid {x}_{t-1}, k_{t}}\, ,\\
         {y}_{t} \sim G^{\theta}\bra{ {y}_{t}\mid  {x}_{t}, k_{t}}\, ,
        \label{eq:model}
\end{gathered}
\end{equation}
\normalsize

\noindent where we have made explicit any dependence on the model parameters $\theta$. $R^{\theta}_{0}$ and $R^{\theta}$ are deterministic functions, $K^{\theta}_0$ and $K^{\theta}$ are categorical distributions. To avoid confusion with the generic SSM Eq. \eqref{eq:SSM}, SSM model components and variables are denoted by a circumflex $\bra{\hat{\ }}$ whereas components of the studied regime switching model, Eq. \eqref{eq:model} are not. For notational simplicity we do not make explicit any time dependence of the model components; by treating the time as a series of constants, time dependence can be introduced without change to the theoretical analysis.

This paper addresses the problem of accurately estimating filtering means, $\mathbb{E}^{\theta}\sbra{ {x}_{t}\mid {y}_{0:t}}$. Unlike previous work \cite{El-Laham2021RSPF, Li2023RSDBPF, Martino2017MMPF}, our formulation allows all of the dynamic model, $M^{\theta}_{0}, \, M^{\theta}$; the observation model, $G^{\theta}$; and the switching dynamic, $R^{\theta}_{0}, \, R^{\theta}, \, K^{\theta}_{0}, \, K^{\theta}$, to depend simultaneously on the learned parameters $\theta$. However, we require that the number of regimes $N_{\text{reg}}$ be given; efficiently determining $N_{\text{reg}}$ is left for future work.
\newline
\textbf{Problem analysis}

The advantage of our formulation is that the joint hidden process $\cbra{{x}_{t}, k_{t},  {r}_{t}}$ is explicitly Markov. Identifying $\cbra{ {x}_{t}, k_{t},  {r}_{t}}$ with $\cbra{ {\hat{x}}_{t}}$ and $\cbra{ {y}_{t}}$ with $\cbra{\hat{y}_{t}}$ it is clear that our model \eqref{eq:model} is a special case of an SSM \eqref{eq:SSM}. Existing results and algorithms concerning SSMs, \ie particle filters, can be applied directly in our setting.

It is instructive to demonstrate a pair of special cases of our model. Taking
\small
\begin{equation}
    R^{\theta}_{0}\bra{k_{0}} = k_{0}, \, R^{\theta}\bra{k_{t},  {r}_{t-1}} = k_{t} \, ,
    \label{eq:MSM}
\end{equation}
\normalsize
one recovers the popular Markov switching model \cite{guidolin2011markovswitching, Blom1985IMM, Mcginnity1998IMM}. Alternatively taking
\small
\begin{equation}
    R^{\theta}_{0}\bra{k_{0}} = \cbra{k_{0}}, \, R^{\theta}\bra{ k_{t},  {r}_{t-1}} =  \bra{\cbra{k_{t}} \cup {r}_{t-1}} \setminus \cbra{k_{t-\tau}} \, ,
    \label{eq:perfect_memory}
\end{equation}
\normalsize
 for some fixed $\tau$, one obtains the a model with perfect memory of its regimes for a fixed lag, similar to the strategy of \cite{chouzenoux2024non-Markov}. Taking $\tau$ to be larger than the trajectory length gives the model with perfect regime memory, as considered by \cite{El-Laham2021RSPF, Li2023RSDBPF}.

 It is well known that, as a consequence of the resampling step, particle filters suffer from path degeneracy \cite{andrieu2005online-estimation}, wherein late-time particles are descended from only a small subset of the early time particles. In the perfect memory formulation this means that late-time particles will form a poor quality sample of the early time regime indices, so keeping full trajectories is not useful. $R^{\theta}$ can be thought of a caching function that keeps the useful information in the regime history. Moreover, myriad practical challenges arise in trying to store and utilise a linearly growing amount of information.

Since the regime may take only one of $N_{\text{reg}}$ values for each time-step, at time $t$, $\abs{\mathcal{R}} \leq N_{\text{reg}}^{t+1}$, with $\abs{\mathcal{R}}$ being the cardinality of the set $\mathcal{R}$. However, an $\mathbb{R}^{d_{r}}$ representation is more convenient for a neural network implementation.

\section{Background}
\label{sec:background}
\subsection{The interacting multiple model particle filter}
\label{sec:IMMPF}
Our algorithm may be seen as differentiable variant of the IMMPF \cite{Blom2007exactIMM}, indeed, we show in Theorem \ref{theorem:weightsnongrad} that the two are exactly equivalent on the forward pass. In \cite{Blom2007exactIMM} the IMMPF is formulated for regime switching systems where the system state and regime-index, $\cbra{x_{t}, k_{t}}$, jointly form a Markov process. Endowing the state with the auxiliary variable $r_{t}$ it is clear that our problem, Eq. \eqref{eq:model}, is a special case of the problem solved by the IMMPF. For clarity, we present the IMMPF as applied to Eq. \eqref{eq:model}.

In filters for regime switching systems, to avoid the sample collapsing to a single regime, it is standard to propose the regime index uniformly \cite{Boers2003IMM, gordon2002model_uncertainty}. The obvious strategy \cite{El-Laham2021RSPF} to filter our system, Eq. \ref{eq:model}, would therefore be to run a particle filter with hidden state $\cbra{x_{t}, k_{t}, r_{t}}$ and observations $y_{t}$. Where $k_{t}$ is proposed uniformly. However, doing so is agnostic to each particles probability of being in its sampled state.

The IMMPF \cite{Blom2007exactIMM} approach is to modify the resampling step as so: for every particle at time $t$, to first choose a regime index, $k_{t}$, then resample a particle at time $t-1$ with weight equal to an estimate of their posterior probability conditional on being propagated to regime $k_{t}$ at time $t$. The other components of the hidden state $\cbra{x_{t}, r_{t}}$ can be propagated in the usual way. Under this approach, the resampling probabilities depend on the value of the latent state at the next time-step, as such it is not possible to express this algorithm as a sequential importance sampling-resampling particle filter \cite{Doucet2009tutorial, chopin2020book}. In practice we deterministically choose there to be an equal number of particles in each regime, but for the sake of exposition we focus on the case that the regime indices are uniformly sampled. We justify this simplification by the observation that the deterministic case can be seen as a deterministic sample of the mixture model resulting from the uniform case, and so is unbiased given the population at the previous time-step \cite{elvira2019genMIS}. 

In Algorithm \ref{alg:SMIS} we introduce a very general sequential Monte-Carlo algorithm that includes both the usual particle filter and the IMMPF as special cases. Consider repeatedly importance sampling a series of target, $\mu\bra{\hat{x}_{t}}$, and proposal, $\lambda_{t}\bra{\hat{x}_{t}}$, distributions that may depend on the sample at the previous time-step. We refer to this algorithm as multiple sequential importance sampling (SMIS), since $\mu_{t\geq1}$ and $\lambda_{t\geq1}$ will be mixtures in all cases of interest.

\begin{algorithm}[h]
\caption{Sequential Multiple Importance Sampling. All operations indexed by $n$ should be repeated for all $n \in \cbra{1, \dots, N}$. $c$ is a, typically unknown, constant.}
\label{alg:SMIS}
\begin{algorithmic}[1]
\REQUIRE proposal mixtures $\lambda_{0}, \lambda_{t}$
\hspace*{1.6em} time extent $T$\\
\hspace*{1.6em} particle count $N$
\vspace*{0.5em}

\STATE ${\hat{x}}^{n}_{0} \sim \lambda_{0}\bra{\hat{x}^{n}_{0}}$;\\
\STATE $w^{n}_{0} \leftarrow \frac{\mu_{0}\bra{\hat{x}^{n}_{0}}}{\lambda_{0}\bra{\hat{x}^{n}_{0}}}$; $\mu_{0}\bra{\hat{x}_{0}}$ is defined in Eq. \eqref{eq:posterior}. \\
\STATE $\bar{w}^{n}_{0} \leftarrow \frac{w^{n}_{0}}{\sum^{N}_{i=1}w^{i}_{0}}$;\\
\vspace{1pt}
\FOR{$t = 1$ \textbf{to} $T$}
    \STATE $\hat{x}^{n}_{t} \sim \lambda_{t} \bra{\hat{x}^{n}_{t}}$
    \STATE $w^{n}_{t} \leftarrow c\frac{\mu_{t}\bra{\hat{x}^{n}_{t}}}{\lambda_{t}\bra{\hat{x}^{n}_{t}}}$; $\mu_{t\geq1}\bra{\hat{x}_{t}}$ is defined in Eq. \eqref{eq:SMIS-target}.\\
    \STATE $\bar{w}^{n}_{t} \leftarrow \frac{w^{n}_{t}}{\sum^{N}_{i=1}w^{i}_{t}}$;\\
\ENDFOR
\RETURN $ {\hat{x}}^{1:N}_{0:T}, \, w^{1:N}_{0:T}, \, \Bar{w}^{1:N}_{0,T}$.
\end{algorithmic}
\end{algorithm}

For filtering, ideally the target is the posterior.
\small
\begin{equation}
\label{eq:posterior}
    \mu_{t}\bra{\hat{x}_{t}} = p\bra{\hat{x}_{t}\mid\hat{y}_{0:t}}\, .
\end{equation}
\normalsize
However, at $t>0$ the true posterior is not typically available so we replace it by an empirical approximation:
\small
\begin{equation}
    \mu_{t\geq1}\bra{\hat{x}_{t}} \defines \frac{\hat{G}\bra{\hat{y}_{t}\mid\hat{x}_{t}}\sum^{N}_{n=1}\bar{w}^{n}_{t-1}\hat{M}\bra{\hat{x}_{t}\mid\hat{x}^{n}_{t-1}}}{p\bra{\hat{y}_{t}\mid\hat{y}_{0:t-1}}}\, .
    \label{eq:SMIS-target}
\end{equation}
\normalsize
We only consider SMIS algorithms with the target distribution given in Eqs. \eqref{eq:posterior} and \eqref{eq:SMIS-target}.

For example, we recover the generic bootstrap particle filter with
\small
\begin{equation}
\label{eq:bootstrap}
    \bra{\lambda_{\text{Boot}}}_{t\geq1} \bra{\hat{x}_{t}} \defines \sum^{N}_{n=1}\bar{w}^{n}_{t-1}\hat{M}\bra{\hat{x}_{t}\mid\hat{x}^{n}_{t-1}}\, ,
\end{equation}
\normalsize
and the IMMPF with uniform index sampling, for model \eqref{eq:model}, with
\small
\begin{multline}
\label{eq:IMMPFprop}
    \bra{\lambda_{\text{IMMPF}}}_{t\geq1}\bra{\hat{x}_{t}} = \bra{\lambda_{\text{IMMPF}}}_{t\geq1}\bra{x_{t}, k_{t}, r_{t}} \\ \defines \frac{\sum^{N}_{n=1} \bar{w}^{n}_{t-1} K^{\theta}\bra{k_{t}\mid r^{n}_{t-1}} M^{\theta}\bra{x_{t}\mid x^{n}_{t-1}, k_{t}}\delta_{R^{\theta}\bra{r^{n}_{t-1}, k_{t}}}\bra{r_{t}}}{N_{\text{reg}}\sum^{N}_{l=1}\bar{w}^{l}_{t-1}K^{\theta}\bra{k_{t}\mid r^{l}_{t-1}}}\, ,
\end{multline}
\normalsize
where $\delta_{z}\bra{\cdot}$ is density of the Dirac measure at $z$. Note that Eq. \eqref{eq:IMMPFprop} cannot be reduced to a mixture density over the particles, motivating the use of the more general SMIS framework.

\begin{theorem}
\label{theorem:Consistency}
Defining $\mathcal{F}_{t}\bra{\psi} \defines \sum^{N}_{n=1}\bar{w}^{n}_{t}\psi\bra{\hat{x}^{n}_{t}}$, and $\mathbb{P}_{t}\bra{\psi}$ to be true posterior mean of some test function $\psi: \mathcal{X} \to \mathbb{R}$, respectively. Then under the sufficient set of assumptions that $\abs{\psi}$ is upper bounded, $\mu_{t}$ is absolutely continuous with respect to $\lambda_{t}$, and the Radon-Nikodym derivative $\frac{\mu_{t}}{\lambda_{t}}$ is finitely upper bounded:
\small
    \begin{equation}
        \mathcal{F}_{t}\bra{\psi} \xrightarrow{L^{2}} \mathbb{P}_{t}\bra{\psi}\, ,
    \end{equation}
    \normalsize
as $N \to \infty$, implying that $\mathcal{F}_{t}\bra{\psi}$ is a weakly consistent estimator of $\mathbb{P}_{t}\bra{\psi}$.
\end{theorem}
\begin{proof}
See \ref{sec:proof-consist} for a proof.
\end{proof}

We are not the first to formalise particle filtering as a SMIS procedure \cite{whiteley2013stability, elvira2018IAPF}, see also \cite{elvira2019auxiliaryMIS} for a tutorial that introduces the well-known auxiliary particle filter \cite{pitt1999auxiliary} as an SMIS procedure. We extend the Theorem 1 to an equivalent result for the deterministic regime sampling case in Corollary \ref{coro:deterministic-case}.

\subsection{Differentiable particle filtering}
A diverse taxonomy of strategies exist to estimate the parameters of SSMs. We refer the readers to \cite{kantas2009estimation} for an overview of SMC approaches. Our problem differs from the classical cases in two respects: using neural networks, our parameter space is very high-dimensional; and with the flexibility we allow for, the latent system is not identifiable from the observations alone. With modern automatic differentiation, the ubiquitous-in-machine-learning, gradient-based schemes are an attractive choice. They perform well when the parameter space is high-dimensional and generalise readily to specialised loss functions. However, standard SMC algorithms are not differentiable.

\emph{Differentiable particle filter} (DPF) refers, in the literature, to any SMC filtering algorithm that is designed to return estimates of the gradients of its outputs. The first DPF \cite{jonschkowski2018DPF} used the well known reparameterisation trick to differentiate sampling from the proposal. But it did not pass gradients through resampling, setting them to zero, so that gradients are not propagated over time-steps.

  In \cite{Corenflos2021OT}, the first fully smoothly differentiable particle filter is proposed. Their strategy is to find a differentiable transport map from the particle approximation of $p(x_{t}\mid y_{0:t-1})$ to $p(x_{t}\mid y_{0:t})$. This results in consistent gradient estimates. However, it is computationally costly and can suffer from numerical issues if not carefully tuned. We refer to this method as \emph{optimal transport resampling} as the map chosen is an approximation of the entropy regularised Wasserstein-2 optimal map.

Several works \cite{naesseth2018variational, le2018auto-encoding, maddison2017variational} have explored applying REINFORCE \cite{williams1992REINFORCE} to differentiate through the discrete resampling steps. However all these papers opted to ignore the gradient due to REINFORCE due to high variance. In \cite{scibior2021stopgrad} it is proposed to apply REINFORCE separately to each resampled particle; however this is less generally applicable than optimal transport resampling. Its gradient estimates have been shown to be consistent \cite{scibior2021stopgrad, arya2022auto-diff} only for a restricted class of filters and functions of the filtering outputs.

The only prior work in differentiable filtering specifically concerned with regime-switching models is the RSDBPF \cite{Li2023RSDBPF}. The RSDBPF is a direct application of the biased DPF of \cite{jonschkowski2018DPF} to the RSPF \cite{El-Laham2021RSPF}. However, it does not attempt to learn the switching dynamic $R^{\theta}, K^{\theta}$ and assumes these components are known a priori.

\section{The Differentiable Interacting Multiple Model Particle Filter}
\label{sec:model}

The core algorithmic contribution of this work is to develop the IMMPF into a differentiable variant, the DIMMPF, so that it can be included into an end-to-end machine learning framework.

\subsection{Parameterising the model}

For the dynamic and observation models, the choice is problem dependent, so we do not make any specific recommendations; the architectures used in our experiments can be found in Section \ref{sec:exp_details}. We take inspiration from long short term memory networks (LSTMs) \cite{sak2014LSTM} to design the parameterisation of $R^{\theta}$. It is well known that SSMs, for which the latent process does not forget its past in some sense result in poorly performing SMC algorithms. Existing bounds on the stability of filtering algorithms typically rely on either conditions that lead to strong mixing \cite{delmoral2004book}, or a related drift condition \cite{whiteley2013stability}. Furthermore, there are SSMs that do not obey these assumptions for which it can be shown that their associated particle filter diverges exponentially (or worse) in mean squared error with $T$ \cite{chopin2020book}.

For this reason, we include forget gates in our parameterisation of the switching dynamic; \ie we set $r_{t} = r_{t-1} \odot a + b$, where every element of vector $a$ is between $0$ and $1$, and $b$ is a function only of $k_{t}$. We desire that this has the effect of decreasing the weight of information from past states at each time-step before introducing information from the noisy $k_{t}$. Algebraically, the model can be expressed as,
\small
\begin{subequations}
\begin{gather}
    \begin{split}
        r_{t} =& R^{\theta}\bra{k_{t}, r_{t-1}} \\
        =& \sigma\bra{\Theta_{1} r_{t - 1}} \odot \sigma \bra{\Theta_{2} k'_{t}} \odot r_{t -1}
        + \text{tanh}\bra{\Theta_{3}  k'_{t}}\, ,
    \end{split}\\
    r_{0} = R^{\theta}_{0}\bra{k_{0}}
        = R^{\theta}\bra{k_{0}, \vec{0}}\, , \\
    K'^{\theta}\bra{k'_{t}\mid r_{t-1}} = \lvert \Theta_{4}\text{tanh}\bra{\Theta_{5}r_{t-1}}\rvert \cdot k'_t\, , \\
        K^{\theta}\bra{k_{t}\mid r_{t-1}} = \frac{K'^{\theta}_{t}\bra{k'_{t}\mid r_{t-1}}}{\sum_{c \in \mathcal{K}} K'^{\theta}\bra{c'\mid r_{t-1}}}\, ,
\end{gather}
\end{subequations}
\normalsize
    where $k'_{t}$ is the one hot encoding of $k_{t}$, $\odot$ is the Hadamard product, and $\Theta_{1:5}$ are learned matrices, $\vec{0}$ is the zero vector, and $\sigma\bra{\cdot}$ is the sigmoid function. $K^{\theta}_{0}$ is represented by a learned vector of regime probabilities.

\begin{figure}
\centering
\begin{tikzpicture}
    \node (inr) [input] {$r_{t-1}$};
    \node (ink) [input, below = 2cm of inr] {$k'_{t}$};
    \node (prod1) [obs, right =  0.5cm of inr] {$\odot$};
    \node (sig1) [cache, above = 0.5cm of prod1] {$\sigma$};
    \node (sig2) [cache, below = 0.8cm of prod1] {$\sigma$};
    \node (plus) [obs, right = 0.5cm of prod1] {$+$};
    \node (tanh1) [cache, below = 0.8cm of plus] {\scriptsize tanh};
    \node (abs) [cache, above right = -0.85 and 0.5cm of tanh1] {\scriptsize abs};
    \node (norm) [obs, below = 0.41cm of abs] {\tiny Normalise};
    \node (tanh2) [cache, above = 0.35cm of abs] {\scriptsize tanh};
    \node (outr) [input, right = 5.5cm of inr] {$r_{t}$}; 
    \node (outk) [input, below = 2cm of outr] {$K^{\theta}$}; 

    \draw[->, thick] (inr.east) -- (prod1.west);
    \draw[->, thick] (sig1.south) -- (prod1.north);
    \draw[->, thick] (inr.east) -- (0.8, 0) |- (sig1.west);
    \draw[->, thick] (prod1.east) -- (plus.west);
    \draw[->, thick] (plus.east) -- (outr.west);
    \draw[->, thick] (ink.east) -| (sig2.south);
    \draw[->, thick] (sig2.north) -- (prod1.south);
    \draw[->, thick] (ink.east) -| (tanh1.south);
    \draw[->, thick] (tanh1.north) -- (plus.south);
    \draw[->, thick] (plus.east) -| (tanh2.north);
    \draw[->, thick] (tanh2.south) -- (abs.north);
    \draw[->, thick] (abs.south) -- (norm.north);
    \draw[->, thick] (norm.east) -- (outk.west);
\end{tikzpicture}
    \caption{Graphical representation of our proposed switching dynamic. Blue nodes are input/outputs. Purple nodes are fully connected network layers with the specified activation. Yellow nodes are non-learned functions. The switching probability mass, $K^{\theta}\bra{k_{t+1}\mid r_{t}}$, is the value at the $k_{t+1}^{\text{th}}$ index of the model output $K^{\theta}$.}
    \label{fig:RSNET}
\end{figure}
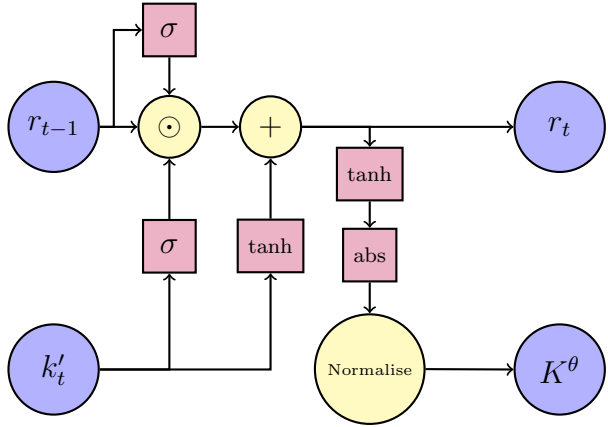

\subsection{Estimating the gradient of the DIMMPF}
\label{sec:Differentiability}
The large majority of DPFs proposed \cite{jonschkowski2018DPF, karkus2018DPF, Corenflos2021OT, chen2024normalisingflow, younis2024differentiable} take derivatives with respect to the proposal distribution's parameters by the low variance reparameterisation trick. However, for our case where the state space has a discrete component no explicit reparameterisation function exists.
Implicit reparameterisations for mixture models \cite{graves2016mixturereparam} have been shown to perform poorly in particle filtering \cite{younis2024differentiable}, so we avoid their usage in favour of REINFORCE. Furthermore, the model index, $k_{t}$, is categorical rather than atomic so continuous relaxations such as the one used in \cite{cox2024gaussMix} do not apply. 

Under the IMMPF proposal it is natural to combine the model selection steps and the resampling steps. We apply REINFORCE separately to each resampled weight as done in \cite{scibior2021stopgrad, arya2022auto-diff} but use a reparameterisation to sample $M^{\theta}\bra{x_{t}\mid x_{t-1}, k_{t}}$,
\small
\begin{subequations}
\label{eq:DIMMPF-weights}
    \begin{equation}
        \tilde{w}^{n}_{t} = \sum^{N}_{m=1}\bar{w}^{m}_{t-1} K^{\theta}\bra{k^{n}_{t}\mid r^{m}_{t-1}} \bot \sbra{M^{\theta}\bra{x^{n}_{t}\mid x^{m}_{t-1},k^{n}_{t}}}\, ,
    \label{eq:DIMMPF-weights-1}
    \end{equation}
    \begin{equation}
        w^{n}_{t} =  \frac{\tilde{w}^{n}_{t}G^{\theta}\bra{y_{t}\mid x^{n}_{t},k^{n}_{t}}\bot\sbra{\sum^{N}_{l=1}\bar{w}^{l}_{t-1}K^{\theta}\bra{k^{n}_{t}\mid r^{l}_{t-1}}}}{N_{\text{reg}}\bot\sbra{ \tilde{w}^{n}_{t}}}\, ,
        \label{eq:DIMMPF-weights-2}
    \end{equation}
\end{subequations}
\normalsize
where we use $\bot\sbra{\cdot}$ to denote the stop gradient operator, which is defined to be the operation that is the identity on the forward pass, but sets the gradient of the enclosed quantity to zero. In modern auto-differentiation libraries, this is easy to implement and computationally cheap by detaching the operand from the computation graph. We refer the reader to \cite{mohamed2020gradientestimation} for an expansive overview of Monte Carlo gradient estimation.
\begin{theorem}
    \label{theorem:weightsnongrad}
    The weighting function Eqs. \eqref{eq:DIMMPF-weights-1} and \eqref{eq:DIMMPF-weights-2}, reduces to the same value as the weights obtained using the $\lambda_{\text{IMMPF}}$, Eq. \eqref{eq:IMMPFprop}, in algorithm \ref{alg:SMIS},  on the forward pass.
\end{theorem}
\begin{proof}
    In the forward pass, Eq. \eqref{eq:DIMMPF-weights-2} simplifies to:
    \small
    \begin{equation}
        N_{\text{reg}}w^{n}_{t} = G^{\theta}\bra{y^{n}_{t}\mid x^{n}_{t},k^{n}_{t}}\sum^{N}_{l=1} \bar{w}^{l}_{t-1}K^{\theta}\bra{k^{n}_{t}\mid r^{l}_{t-1}}\, .
        \label{eq:simplified-weights}
    \end{equation}
    \normalsize
    Evaluating Eq. \eqref{eq:IMMPFprop} for a given latent state gives:
    \small
    \begin{equation}
        \bra{\lambda_{\text{IMMPF}}}\bra{x^{n}_{t}, k^{n}_{t}, r^{n}_{t}} = \frac{\sum^{N}_{m=1}\bar{w}^{m}_{t-1}K^{\theta}\bra{k^{n}_{t}\mid r^{m}_{t-1}}M^{\theta}\bra{x^{n}_{t}\mid x^{m}_{t-1}, k^{n}_{t}}}{N_{\text{reg}}\sum^{N}_{l=1}\bar{w}^{l}_{t-1}K^{\theta}\bra{k^{n}_{t}\mid r^{l}_{t-1}}}\, .
        \label{eq:eval-prop}
    \end{equation}
    \normalsize
    Then calculating $w^{t}_{n}$ as in line 6 of Algorithm \ref{alg:SMIS} from Eq. \eqref{eq:eval-prop}, gives Eq. \eqref{eq:simplified-weights}. 
\end{proof}
\begin{theorem}
\label{theorem:consistency-DIMMPF}
     Under the weights given in Eq. \eqref{eq:DIMMPF-weights};  assuming the model components are such that the proposal dominates the target; and the magnitudes of $w^{n}_{t}$, $\grad w^{n}_{t}$, $\psi$ and $\grad\psi$ are upper-bounded for all $t$ then,
    \begin{equation}
        \grad\mathcal{F}_{t}\bra{\psi} \xrightarrow{L^{2}} \grad \mathbb{P}_{t}\bra{\psi}
    \end{equation}
    as $N \to \infty$, implying that $\grad\mathcal{F}_{t}\bra{\psi}$ is a weakly consistent estimator of $\grad\mathbb{P}_{t}\bra{\psi}$.
\end{theorem}
\begin{proof}
    See \ref{sec:proof-DIMMPF} for a proof.
\end{proof}

Our weight function, Eq. \eqref{eq:DIMMPF-weights}, is not the unique estimator that applies REINFORCE separately to each resampled particle. We improve on the na\"{i}ve choice,
\begin{equation}
\label{eq:Naive-weights}
    \bra{w_{\text{\tiny Na\"{i}ve}}}^{n}_{t} = G^{\theta}\bra{y_t\mid x^{n}_{t}, k^{n}_{t}}\frac{\bar{w}^{a^{n}_{t}}_{t-1}}{\bot\sbra{\bar{w}^{a^{n}_{t}}_{t-1}}}\bot\sbra{\sum^{N}_{l=1}\bar{w}^{l}_{t-1}K^{\theta}\bra{k^{n}_{t}\mid r^{l}_{t-1}}} \, ,
\end{equation}
where $a^{n}_{t}$ is the index of the particle at time $t-1$ that particle $n$ at time $t$ has been propegated from, by, at each time-step, partially Rao-Blackwellising over the previous time-step. Only the REINFORCE gradient terms benefit from this Rao-Blackwellisation, reparameterised gradients track only through the ancestral path. We find experimentally, in Table \ref{tab:results}, that the Rao-Blackwellised estimator performs significantly better than its na\"{i}ve counterpart. The na\"{i}ve estimator's variance is such that it performs worse than the biased estimator that zeros out the REINFORCE terms.

The Rao-Blackwellistaion comes at a computational cost; the density $M^{\theta}\bra{x^{n}_{t}|x^{m}_{t-1},k^{n}_{t}}$ must be computed for every pair $\cbra{n,m}$ at an operation cost of $\mathcal{O}\bra{N^{2}}$ in the forward pass compared to the usual particle filter $\mathcal{O}\bra{N}$ complexity. In practice these operations can be computed in parallel so the Rao-Blackwellised estimator, Eq. \eqref{eq:DIMMPF-weights}, is only slightly slower to compute than the na\"{i}ve estimator, Eq. \eqref{eq:Naive-weights}. Furthermore, by Theorem 2, the Rao-Blackwellisation only affects the gradient estimators, so we revert to the $\mathcal{O}\bra{N}$ algorithm during inference. See Table \ref{tab:time} for our results.

We remark that the proof of Theorem \ref{theorem:consistency-DIMMPF}, and its corollaries rely on the fact that, using the DIMMPF weights Eq. \eqref{eq:DIMMPF-weights}, gradients are only taken with respect to the model components and not the proposal. For this reason, one cannot expect our estimator to provide a useful learning signal for the proposal process. In \cite{scibior2021stopgrad} the authors investigate a closely related estimator and are unable to find clear theoretical or experimental evidence that it does. Under our set-up the proposal has no extra parameters in addition to the model. Developing a practical gradient estimator that can learn an efficient proposal process remains an open problem.

\subsection{Training the DIMMPF}
Consider two approaches to train a particle filter when there is access to the ground truth latent state during training. The first is to estimate the latent state and minimise some distance between the estimator and the ground truth. The second is to maximise the joint likelihood of the observations and their associated ground truth latent state. We find the best results are obtained when optimising a loss that combined the two strategies. A related strategy is recommended in \cite{jonschkowski2018DPF} where the optimisation objective is a combination of the MSE of filtering estimates and a loss on the measurement and observation models individually.

In our case, the MSE of filtering estimates is obtained by Algorithm \ref{alg:DIMMPF}. To estimate the joint likelihood, we re-partition the model so that during training the available quantities $\cbra{x_{t}, y_{t}}$ are taken as observations and $\cbra{r_{t}, k_{t}}$ are the latent state estimated by filtering. Since the observations now depend on each other, this formulation does not strictly satisfy the usual definition of an SSM, described in Eq. \eqref{eq:SSM}. However, in particle filtering, the observations are treated as a sequence of non-random constants so the algorithm generalises freely to situations where the observations have arbitrary backwards-in-time interdependence.
To account for this interdependence, we replace the conditional likelihood in Eq. \eqref{eq:SSM}, with 
\begin{equation}
    \hat{y}_{t} \sim \hat{G}\bra{\hat{y}_{t}\mid\hat{x}_{t}, \hat{y}_{0:t-1}}\, .
\end{equation}
We provide precise details on how we calculate these losses in \ref{app:losses}.

Needing to run two filters incurs an extra computational cost. However, for the second filter the conditional likelihood only depends on the state through the discrete model index, so one can precompute all the conditional likelihoods for each choice of model index using GPU parallelism. Then the only neural network we need to evaluate per-particle during filtering is $R^{\theta}\bra{k_{t}, r_{t-1}}$.

Formally, Theorem \ref{theorem:consistency-DIMMPF} does not prove the consistency of the gradients of either the losses we adopt, however the appropriate results follow as corollaries. See Corollaries \ref{coro:MSE-consist} and \ref{coro:likelihood-consist} for formal statements and proofs of the consistency of the gradients of the MSE and log-likelihood respectively. When we calculate the log-likelihood of $\cbra{x_{t}, y_{t}}$ using the re-partioned model, the state dynamics have no reparameterised component. In \cite{scibior2021stopgrad} it is shown, for this case, that the resultant gradient estimator is unbiased. It is closely related to the Rao-Blackwellised recursion obtained in \cite{poyiadjis2011fishers}.

\subsection{The full algorithm and practical considerations}
We present the full DIMMPF algorithm\footnote{Our implementation, including the code to run the experiments in this paper can be found at https://github.com/John-JoB/DIMMPF.} in pseudo-code in Algorithm \ref{alg:DIMMPF}. It can be checked that the filtering loop of Algorithm \ref{alg:DIMMPF} is equivalent, in the forward pass, to running Algorithm \ref{alg:SMIS} with target \eqref{eq:SMIS-target} and proposal \eqref{eq:IMMPFprop}, with the model indices deterministically chosen rather than sampled uniformly.
When computing the data-likelihood, Algorithm \ref{alg:DIMMPF} is modified such that the distribution of $x_{t}$ is accounted for in the weights instead of being sampled from, see \ref{app:losses} for detail.

In practice, it is common to use a variance-reduced scheme to resample the particle indices in lieu of multinomial sampling. One such scheme is the systematic resampling of \cite{carpenter1999sytematic} and its closely related variants, which are found to work well empirically. The increased stability offered by systematic resampling is of increased importance in the context of DPFs, where it stabilises gradient updates as well as the forward pass \cite{zhu2020diff-resample}. However, under systematic resampling the particles are no longer sampled i.i.d., so the central limit theorem used to prove Theorem \ref{theorem:Consistency} no longer holds. We make the recommendation that practitioners use systematic resampling based on our empirical results.

\begin{algorithm}[H]
\caption{Differentiable Interacting Multiple Model Particle Filter. All operations indexed by $n$ should be repeated for all $n \in \cbra{1, \dots, N}$ and those by $q$ for $q \in \cbra{1, \dots, N_{\text{reg}}}$.}
\label{alg:DIMMPF}
\begin{algorithmic}[1]
\REQUIRE priors $M^{\theta}_{0}$
\hspace*{6.05em} dynamic models $M^{\theta}$\\
\hspace*{1.5em} regime prior $K^{\theta}_{0}$
\hspace*{3.5em} switching dynamic $K^{\theta}$\\
\hspace*{1.5em} observation models $G^{\theta}$
\hspace*{0.7em} encoding functions $R^{\theta}$ \\
\hspace*{1.5em} time length $T$
\hspace*{4.5em} particle count $N$ \\
\hspace{1.5em} loss coefficient $\lambda$
\hspace*{3.3em} observations $y_{0:T}$ \\
\hspace*{1.5em} ground truth $\tilde{x}_{0:T}$
\hspace*{2.8em} number of regimes $N_{\text{reg}}$
\ENSURE unormalised weights $w^{1:N}_{0:T}$
\hspace*{1em} particle locations $\cbra{x^{1:N}_{0:T}, k^{1:N}_{0:T}, r^{1:N}_{0:T}}$
\STATE $k^{n}_{0} \leftarrow \lfloor \frac{nN_{\text{reg}}}{N} \rfloor$
\STATE Sample $x^{n}_{0} \sim M^{\theta}_{0}\bra{x^{n}_{0}\mid k^{n}_{0}}$
\STATE $r^{n}_{0} = R^{\theta}\bra{k^{n}_{0}, \vec{0}}$
\STATE $w^{n}_{0} = G^{\theta}\bra{y_{0}\mid x^{n}_{0}, k^{n}_{0}}$
\STATE $\bar{w}^{n}_{0} = \frac{w^{n}_{0}}{\sum^{N}_{m=1}{w^{m}_{0}}}$
\FOR{$t= 1 \text{ to } T$}
    \STATE $k^{n}_{t} \leftarrow \lfloor \frac{nN_{\text{reg}}}{N} \rfloor$
    \STATE $\hat{w}^{n,q}_{t} \leftarrow \frac{\bar{w}^{n}_{t-1}K^{\theta}\bra{q | r^{n}_{t-1}}}{\sum^{N}_{m=1} \bar{w}^{m}_{t-1}K^{\theta}\bra{q | r^{m}_{t-1}}}$
    \STATE Sample ancestor indices $a^{n}_{t} = m$ with probability equal to $\hat{w}^{m,k^{n}_{t}}_{t}$
    \STATE Sample $x^{n}_{t} \sim M^{\theta}\bra{x^{n}_{t}\mid x^{a^{n}_{t}}_{t-1}, k^{n}_{t}}$
    \STATE $r^{n}_{t} \leftarrow R^{\theta}\bra{k^{n}_{t}, r^{a^{n}_{t}}_{t-1}}$
    \STATE Calculate $w^{n}_{t}$ from Eq. \eqref{eq:DIMMPF-weights}
    \STATE $\bar{w}_{t}^{n} = \frac{w_{t}^{n}}{\sum^{N}_{m=1}{w_{t}^{m}}}$
\ENDFOR
\vspace{0.2em}
\RETURN $w^{1:N}_{0:T}, x^{1:N}_{0:T}, k^{1:N}_{0:T}, r^{1:N}_{0:T}$
\end{algorithmic}
\end{algorithm}

\section{Numerical Experiments}
\label{sec:experiments}
In this section, we present the results from a set of numerical experiments.

\subsection{Simulated environments}
We repeat the test environment of \cite{El-Laham2021RSPF, Li2023RSDBPF}, in which the dynamic and observation models of each regime are uni-variate and Gaussian.
\small
\begin{subequations}
    \begin{gather}
        M_0\bra{x_0} = \mathcal{U}\bra{-0.5, 0.5}\,, \\
        M\bra{x_{t}| x_{t-1}, k_{t}} = \mathcal{N}\bra{a_{k_{t}} x_{t-1} + b_{k_{t}}, \sigma^{2}}\,, \\
        G\bra{y_{t}| x_{t}, k_{t}} = \mathcal{N}\bra{a_{k_{t}} \sqrt{\lvert x_{t} \rvert} + b_{k_{t}}, \sigma^{2}}\,, \\
        \sbra{a_{1}, \dots, a_{8}} = \sbra{-0.1, -0.3, -0.5, -0.9, 0.1, 0.3, 0.5, 0.9}\,, \\
        \sbra{b_{1}, \dots, b_{8}} = \sbra{0, -2, 2, -4, 0, 2, -2, 4}\,, \\
        \sigma^{2} = 0.1\,.
    \end{gather}
\end{subequations}
\normalsize
This model poses some challenges to state estimation. Because the observation location depends on the state only through its absolute value, it is impossible to estimate the state using the observations alone. Furthermore, the coefficients $a_{i}, b_{i}$ are chosen so that it is hard to identify the current regime over short sequences, for example, regimes $1 \text{ and }5$ have identical data-likelihoods when each is run in isolation. It is therefore required that all of the observation model, the dynamic model and the switching dynamic are well-learned.

We include three different switching dynamics. The first is a Markov switching system where the probability of remaining in the same regime is $0.8$; switching to the next regime, with regimes $9$ and $1$ identified, is $0.15$; and all other regimes have probability $\frac{1}{120}$. Algebraically:
\small
\begin{subequations}
    \begin{gather}
        K\bra{k_{t}|k_{0:t-1}} = \bra{\mathbf{k'}_{t-1}}^{T} B \mathbf{k'}_{t}\,, \\
    B = 
    \begin{pmatrix}
        0.8 & 0.15 & \rho & \dots & \rho \\
        \rho & 0.8 & 0.15 & \dots & \rho \\
        \vdots & & \ddots & & \vdots \\
        \rho & \rho & \dots & 0.8 & 0.15 \\
        0.15 & \rho & \dots & \rho & 0.8 
    \end{pmatrix}\,,\\
    \rho = \frac{1}{120}\,,
    \end{gather}
\end{subequations}
\normalsize
where $k'_{t}$ are the one-hot encodings of the regime index.

In the second setting, the regimes follow a P\'{o}lya-urn distribution where the sampled regimes are more likely to appear at later time-steps:
\small
\begin{equation}
    K\bra{k_{t}|k_{0:t-1}} = \frac{ 1 + \sum^{t-1}_{s=0}\mathds{1}\bra{k_{s} = k_{t}}}{8 + t}\,.
\end{equation}
\normalsize
The P\'{o}lya-urn setting has frequent switching and often the distribution of model indices looks close to uniform, making it simpler to approximate but harder to perform inference on than the Markov setting.
In both cases we set $K_{0}\bra{k_{0}} = \frac{1}{8}$.

To demonstrate the versatility of our algorithm, we introduce a more challenging switching dynamic than has been used in previous work. For the third setting, the time between regime switches is approximately Erlang distributed. The order of the Erlang distribution is equal to the number of periods for which the system has been in the current regime. Once the Erlang distributed period is finished, the system jumps randomly to one of the two adjacent regimes. However, we include a small probability that at any time-step the system jumps to any regime. This system is most simply expressed algebraically as
\small
\begin{subequations}
    \begin{gather}
        m_{t} \sim \text{Bernoulli}\bra{0.01} \,, \\
        n_{t} \sim \text{Bernoulli}\bra{0.2} \,,\\
        c\bra{k_{0:t}, k} = \sum^{t-1}_{s=0}\mathds{1}\bra{\bra{k_{s} = k} \land \bra{\neg \bra{k_{s+1} = k} \lor \bra{m_s = 1}}} \,,\\
        l_{t} = \begin{cases}
            l_{t-1}, \;\; n_{t} = 0\,, \\
            l_{t-1} - 1,  \;\; \neg \bra{l_{t-1} = 0} \land \bra{n_{t} = 1} \, , \\
            c\bra{k_{0:t}, k_{t}}, \; \; \bra{l_{t-1} = 0} \land \bra{n_{t} = 1} \,, 
        \end{cases} \\
        \alpha_{t} = \bra{n_t = 1} \land \bra{l_{t-1} = 0} \, ,\\
        K\bra{k_{t}|k_{0:t-1}} = \begin{cases}
        \frac{1}{N_{\text{reg}}}, \; \; m_t = 1 \, ,\\
        \mathds{1}\bra{k_{t} = k_{t-1}}, \; \;  \bra{m_t = 0} \land \neg \alpha_{t} \, ,\\
        \bra{0.6\mathds{1}\bra{k_{t} = k_{t-1} + 1 \bra{\text{mod} \; N_{\text{reg}}}}\atop+ 0.4\mathds{1}\bra{k_{t} = k_{t-1} - 1 \bra{\text{mod} \; N_{\text{reg}}}}}, \; \; \bra{m_t = 0} \land \alpha_{t}\, .
        \end{cases}
    \end{gather}
\end{subequations}
\normalsize
We choose this dynamic because of its complexity to learn. There are both strong dependence between the index at successive time-steps, like the Markov setting; and long term dependencies, like the P\'{o}lya setting.

\subsection{Experiment details}
\label{sec:exp_details}
In addition to the our DIMMPF, we present a number of baseline approaches. The problem of sequential state estimation can be described as learning to predict a sequence of latent states from a sequence of observations, so any available sequence-to-sequence techniques can apply. We choose to use a transformer\cite{Vaswani2017Transformer} and an LSTM \cite{sak2014LSTM} as baseline approaches as they represent the state-of-the-art in sequence-to-sequence prediction. The transformer is encoder only and the LSTM is unidirectional so that only past information is used. We also compare to the regime learning particle filter (RLPF), a preliminary version of our methodology that we presented in the conference paper \cite{brady2024RLPF}. Finally we include two variants on the DIMMPF: the DIMMPF-OT that uses a transport map based resampler \cite{Corenflos2021OT} to be differentiable, instead of the gradient estimator developed in Section \ref{sec:Differentiability}; and the DIMMPF-N that uses the $\mathcal{O}\bra{N}$ na\"{i}ve gradient estimator, Eq. \eqref{eq:Naive-weights}.

All filtering based models parameterise both the measurement and dynamic models with fully connected neural networks of two hidden layers containing $11$ nodes each. During training we use a population of $200$ total particles, which we increase to $2000$ for testing. This is reduced to $80$ particles in training and $800$ in testing for the DIMMPF-OT due to memory constraints. We generate $2000$ trajectories of $51$ time-steps and use them in ratio $2:1:1$ for training, validation and testing, respectively. We train in mini-batches of $100$ trajectories, but test on the full $500$ trajectory batches. Each experiment is repeated $20$ times with independent data generations. All experiments are performed using an NVIDA RTX 3090 GPU. 

\begin{table}[H]
\small
    \centering
    \caption{Filtering accuracy for the discussed algorithms. Reported values are the achieved mean squared filtering error and averaged across 20 independent training runs.}
    \begin{tabular}{||c|c|c|c||}
    \hline
        Algorithm & Markov MSE & P\'{o}lya MSE & Erlang \\
    \hline
    \hline
    Transformer (baseline) &$1.579\pm0.169$&$1.508\pm0.112$& $1.614 \pm 0.160$\\
     \hline
     LSTM (baseline) &$0.732\pm0.083$ & $0.667\pm0.053$ & $0.978 \pm 0.103$\\
     \hline
     RLPF (baseline) &$0.536\pm0.143$&$0.509\pm0.071$ & $0.771 \pm 0.110$\\
     \hline
     DIMMPF-OT (baseline) & $0.891 \pm 0.128$ & $0.866 \pm 0.134$  & $0.873 \pm 0.122$\\
     \hline
     DIMMPF-N (baseline) & $0.751 \pm 0.0694$ & $0.741 \pm 0.071$ & $0.742 \pm 0.072$ \\
     \hline
     DIMMPF (ours) &$\mathbf{0.500 \pm 0.100}$&  $\mathbf{0.490 \pm 0.052}$ & $\mathbf{0.712 \pm 0.115}$\\
     \hline
     IMMPF (oracle) &$0.274\pm0.019$&$0.408\pm0.014$ & $0.473 \pm 0.025$\\
     \hline
    \end{tabular}
    \label{tab:results}
\end{table}

\begin{table}[H]
\small
    \centering
    \caption{Average computation times per training epoch (10 batches of 100 parallel filters of 200 particles each) and testing run (1 batch of 500 parallel filters of 2000 particles each) on the P\'{o}lya experiment.}
    \begin{tabular}{||c|c|c||}
    \hline
        Algorithm & Av. train epoch time (s) & Av. test time (s) \\
    \hline
    \hline
    Transformer (baseline) &$0.182$&$0.00310$\\
     \hline
     LSTM (baseline) &$\mathbf{0.0145}$&$\mathbf{0.000792}$\\
     \hline
     RLPF (baseline) &$5.98$ & $0.814$\\
     \hline
     DIMMPF-OT (baseline) & $425$ & Out of memory \\
     \hline
     DIMMPF-N (baseline) & $8.56$ & $0.773$  \\
     \hline
     DIMMPF (Ours) &$10.5$ & $0.759$ \\
     \hline
    \end{tabular}
    \label{tab:time}
\end{table}

\subsection{Results}
We present the main results in Table \ref{tab:results}, and the computation times in Table \ref{tab:time}. The DIMMPF is the best performing algorithm in all experiments. The filtering approaches far outperform the generic sequence-to-sequence techniques in mean accuracy, however, the LSTM is computationally the cheapest. In training, the DIMMPF is faster than the DIMMPF-OT. But, it is slower than the DIMMPF-N due requiring more terms to be computed. The RLPF further saves time through ignoring gradient terms that the DIMMPF and DIMMPF-N evaluate. During inference, the DIMMPF and DIMMPF-N are equivalent so achieve similar timings.

\section{Conclusions}
\label{sec:conclusions}
In this paper, we have presented a novel differentiable particle filter, the DIMMPF, that addresses the problem of learning to estimate the state of a regime switching state space process. Our algorithm improves over the previous state-of-the-art, the RSDBPF, in three respects. Firstly, the RSDBPF required that the switching dynamic be fully known a priori, whereas our algorithm can learn it from data. Secondly, the DIMMPF takes account of assigned regime when resampling particles, thereby concentrating computation on more promising regions. Thirdly, the gradient estimates returned by the DIMMPF are consistent.

We evaluated our algorithm on a set of numerical experiments. The three settings, Markov, P\'{o}lya, and Erlang are designed to test the learning of short-term strong dependency, long-range weak dependency, and both simultaneously, respectively. The proposed DIMMPF leads to the smallest filtering errors on all three settings. The DIMMPF is computationally expensive during training, both compared to its simpler variants and especially out-of-the-box sequence-to-sequence techniques. However, during inference it achieves a similar speed to the other DPF approaches.

The key limitations of this work are that the DIMMPF is only capable of learning model parameters and not, additionally, an efficient proposal process; and that the number of regimes must be known. We leave addressing these limitations to future work.

Another important direction for future work is towards more challenging environments, including real-world data. We propose a simple architecture to parameterise the switching dynamic; future work might consider more advanced design patterns such as attention.

\section{CRediT author statement}
\textbf{John-Joseph Brady:} Conceptualisation, Methodology, Software, Formal Analysis, Writing - Original Draft, Writing - Review \& Editing. \textbf{Yuhui Luo:} Supervision, Writing - Review \& Editing. \textbf{Wenwu Wang:} Supervision, Writing - Review \& Editing. \textbf{V\'{i}ctor Elvira:} Writing - Review \& Editing. \textbf{Yunpeng Li:} Conceptualisation, Writing - Review \& Editing, Supervision.

\section{Acknowledgements}
The authors acknowledge funding from the UK Department for Science, Innovation and Technology through the 2024 National Measurement System programme. JJ. Brady was supported in this work through the National Physical Laboratory and University of Surrey partnership via an Engineering and Physical Sciences Research Council studentship. The work of V. E. is supported by ARL/ARO under grant W911NF-22-1-0235. 

\appendix
\section{Calculating the losses}
\label{app:losses}
\textbf{Mean squared error loss:} The mean squared error of the filtering estimates is calculated as follows, with $\tilde{x}_{t}$ denoting the ground truth latent state.
\small
\begin{equation}
\label{eq:MSE}
    \mathcal{L}_{\text{MSE}} = \frac{1}{T+1}\sum^{T}_{t=0}\bra{\sum^{N}_{n=1}\bar{w}^n_{t}\bra{x^{n}_{t} - \tilde{x}_{t}}}^{2} \,.
\end{equation}
\normalsize
\textbf{Data-likelihood loss:} Calculating the data-likelihood requires a modification to the inputs of Algorithm \ref{alg:DIMMPF}. We replace the observation model, $G^{\theta}\bra{y_{t}\mid x^{n}_{t}, k^{n}_{t}}$, with the marginal data-likelihood conditioned on the previous time-step,
\small
\begin{equation}
    G^{\theta}\bra{y_{t}\mid \tilde{x}_{t}, k^{n}_{t}}M^{\theta}\bra{\tilde{x}_{t}\mid\tilde{x}_{t-1}, k^{n}_{t}}.
\end{equation}
\normalsize
In practice we do not track a latent state $x_{t}$, but the same effect can be achieved in Algorithm \ref{alg:DIMMPF}, by setting $M^{\theta}\bra{x_{t}|x_{t-1}, k_{t}}$ to be uniform over some set that does not depend on $x_{t-1}$ or $k_{t}$. Then the log data-likelihood is estimated as
\small
\begin{equation}
\label{eq:likelihood-loss}
    -\mathcal{L}_{\text{data-likelihood}} = \sum^{T}_{t=0}\log\sum^{N}_{n=1}w^{n}_{T} - \bra{T+1}\log{N} \,.
\end{equation}
\normalsize
\section{Proof of Theorem \ref{theorem:Consistency}}
\label{sec:proof-consist}
    The following proof is based, in part, on the proofs found in Chapters 8 and 11 of \cite{chopin2020book}.
    \scriptsize
    \begin{multline}
            \mathbb{E}\sbra{\bra{\mathcal{F}_{t}\bra{\psi} - \mathbb{P}_{t}\bra{\psi}}^{2}} =
            \mathbb{E}\sbra{\bra{\mathcal{F}_{t}\bra{\psi} - \int_{\mathcal{X}}\mu_{t}\bra{d\hat{x}_{t}}\psi\bra{\hat{x}_{t}} + \int_{\mathcal{X}}\mu_{t}\bra{d\hat{x}_{t}}\psi\bra{\hat{x}_{t}} - \mathbb{P}_{t}\bra{\psi}}^{2}} \\ 
            \leq 2\cbra{\mathbb{E}\sbra{\bra{\mathcal{F}_{t}\bra{\psi} - \int_{\mathcal{X}}\mu_{t}\bra{d\hat{x}_{t}}\psi\bra{\hat{x}_{t}}}^{2}}
            + \mathbb{E}\sbra{\bra{\int_{\mathcal{X}}\mu_{t}\bra{d\hat{x}_{t}}\psi\bra{\hat{x}_{t}} - \mathbb{P}_{t}\bra{\psi}}^{2}}}\,.
            \label{eq:MSE_conv}
    \end{multline}
    \normalsize
    The first term can be bounded using the standard convergence result for an auto-normalised importance sampler. For some factor $c_{t}$ that is independent of $N$,
    \footnotesize
    \begin{equation}
    \label{eq:auto-normalised-conv}
    \begin{split}
        &\mathbb{E}\sbra{\bra{\mathcal{F}_{t}\bra{\psi} - \int_{\mathcal{X}}\mu_{t}\bra{d\hat{x}_{t}}\psi\bra{\hat{x}_{t}}}^{2}} \\ 
        &\leq \mathbb{E}\sbra{\norm{\bra{\mathcal{F}_{t}\bra{\psi} - \int_{\mathcal{X}}\mu_{t}\bra{d\hat{x}_{t}}\psi\bra{\hat{x}_{t}}}}^{2}_{\infty}\mid\hat{x}^{1:N}_{t-1}, w^{1:N}_{t-1}} \leq  \frac{1}{N}c_{t}\, ,
    \end{split}
    \end{equation}
    \normalsize
     where we have used the fact that the particles are conditionally i.i.d. given the particles at the previous time-step.
    For $t=0$, the second term in Eq. (\ref{eq:MSE_conv}) is zero, so MSE convergence is guaranteed \ie Theorem \ref{theorem:Consistency} holds for $t=0$. For $t>0$, we may bound the MSE by induction with $t=0$ as the base case. To perform the inductive step we first introduce the identity:
    \footnotesize
    \begin{equation}
        \mathbb{P}_{t-1}\bra{\int_{\mathcal{X}}\psi\bra{\hat{x}_{t}}p\bra{d\hat{x}_{t}\mid \hat{x}_{t-1},\hat{y}_{t}}} = \mathbb{P}_{t}\bra{\psi}\, ,
        \label{eq:propogate}
    \end{equation}
    \normalsize
    which may be proved using basic probability rules and the conditional independence structure of the SSM (Eq. \eqref{eq:SSM}). One may write:
    \footnotesize
    \begin{equation}
    \begin{split}
        \int_{\mathcal{X}}\mu_{t}\bra{d\hat{x}_{t}}\psi\bra{\hat{x}_{t}} &= \sum^{N}_{i=1}\bar{w}^{i}_{t-1}\int_{\mathcal{X}}\psi\bra{\hat{x}_{t}}p\bra{d\hat{x}_{t} \mid \hat{x}^{i}_{t-1}, \hat{y}_{0:t}} \\
        &= \mathcal{F}_{t-1}\bra{\int_{\mathcal{X}}\psi\bra{\hat{x}_{t}}p\bra{d\hat{x}_{t}\mid \hat{x}_{t-1},\hat{y}_{t}}}\, .
    \end{split}
    \end{equation}
    \normalsize
    This implies, using Eq. \eqref{eq:propogate}, that the second term in \eqref{eq:MSE_conv} is bounded by a term of order $N^{-1}$ assuming that Theorem \ref{theorem:Consistency} holds for $t-1$. Then, the MSE becomes the sum of two bounded by order $N^{-1}$ terms. So, it is bounded by $\frac{1}{N}c^{'}_t$ for some constant $c^{'}_{t}$ and therefore the MSE converges to zero in the large sample size limit.

\begin{corollary}
\label{coro:deterministic-case}
The IMMPF with deterministic sampling exhibits $L^{2}$ convergence under the same conditions as Theorem 1,
    \begin{equation}
        \bra{\mathcal{F}_{\text{\tiny Det}}}_{t}\bra{\psi} \xrightarrow{L^{2}} \mathbb{P}_{t}\bra{\psi}\, ,
    \end{equation}
    implying the weak consistency of $\bra{\mathcal{F}_{\text{\tiny Det}}}_{t}\bra{\psi}$.
\end{corollary}

    In the deterministic case the particles are no longer i.i.d given the previous particles, so we can no longer apply the convergence result for the auto-normalised importance sampler to obtain Eq. (A.2).

The IMMPF sampling algorithm is equivalent to scheme N1 in \cite{elvira2019genMIS} where each sampling distribution in repeated for $\frac{N}{N_{\text{reg}}}$ copies. In \cite{elvira2019genMIS} Appendix C Option 3 implies unbiasedness and Equation (D.8) gives the variance of the unnormalised sum of weighted samples for a zero mean quantity,
\small
\begin{equation}
    \hat{\psi} \defines \sum^{N}_{n=1}w\bra{x^{n}}\psi\bra{x^{n}} \, ,
\end{equation}
\normalsize
to be
\small
\begin{equation}
    \var{\hat{\psi}} = \frac{N}{N_{\text{reg}}}\sum^{N_{\text{reg}}}_{k=1}\eE{\bra{w\bra{x}\psi\bra{x}}^{2}|k} \leq \norm{w}_{\infty}^{2}\norm{\psi}_{\infty}^{2}.
\end{equation}
\normalsize
Then by Slutsky's theorem, the asymptotic variance of the proceeding auto-normalised importance sampler is
\small
\begin{equation}
\label{eq:asym-var-det}
    \var{\frac{\hat{\psi}}{\sum^{N}_{m=1}w\bra{x^{m}}}} \to \frac{\eE{\bra{w\bra{x^k}\psi\bra{x^{k}}}^{2}}}{N\eE{w\bra{x}}^{2}} \, .
\end{equation}
\normalsize
Which is exactly the same asymptotic variance one obtains using uniform sampling. This convergence result can be substituted for that of the usual auto-normalised importance sampler in (A.2) and the rest of the proof is identical to that of Theorem 1.

We remark that although the asymptotic variances of the auto-normalised importance samplers for the deterministic, Eq. (A.8), and uniform cases are identical, this does not imply the variances of the filtering estimates are.

\section{Proof of Theorem \ref{theorem:consistency-DIMMPF}}
\label{sec:proof-DIMMPF}
This result and the attached corollaries apply to the deterministic sampling case by applying Corollary \ref{coro:deterministic-case} wherever their proofs use Theorem \ref{theorem:Consistency} for the uniform case.
Throughout this derivation, if the distribution an expectation is taken over is not specified it may be assumed to be the full posterior $p\bra{x_{T},k_{T},r_{T}|y_{0:T}}$. 

Consider generating $N \to \infty$ trajectories from the bootstrap particle filter with, Eq. \eqref{eq:bootstrap}, for our system, Eq. \eqref{eq:model}, with $M^{\theta}\bra{x_{t}|x_{t-1}}$ sampled by the reparameterisation trick. Using the surrogate objectives of \cite{schulman2015gradientgraph}, we can directly write the appropriate gradient estimator for this stochastic program as:
\footnotesize
\begin{multline}
    \grad\underset{\text{\tiny Boot}}{\mathbb{E}}\sbra{\sum^{N}_{n=1}\Omega^{n}_{T}\psi\bra{x^{n}_{T}}} = \underset{\text{\tiny Boot}}{\mathbb{E}}\Biggl[\grad\sum^{N}_{n=1}\Omega^{n}_{T}\psi\bra{x^{n}_{T}} \\ + \sum^{N}_{n=1}\Omega^{n}_{T}\psi\bra{x^{n}_{t}}\grad\sum^{N}_{m=1}\bra{\sum^{T-1}_{t=0}\sbra{\log\Omega^{m}_{t}+\log K^{\theta}\bra{k^{m}_{t+1}|r^{a^{m}_{t}}_{t}}} +  \log K^{\theta}_{0}\bra{k^{m}_{0}}} \Biggr] \, ,
\end{multline}
\normalsize
for ancestor variables $a^{0:N}_{0:T-1}$ such that $a^{n}_{t}$ is the index at time $t$ that particle $n$ at time $t+1$ descends from and
\begin{equation}
    \Omega^{n}_{t} \defines \frac{G^{\theta}\bra{y_{t}|x^{n}_{t},k^{n}_{t}}}{\sum^{N}_{l=1}G^{\theta}\bra{y_{t}|x^{l}_{t},k^{l}_{t}}} \, .
\end{equation}
Taking the limit $N\to \infty$ and applying the consistency result of Theorem \ref{theorem:Consistency} then:
\footnotesize
\begin{multline}
\label{eq:asym-gradient}
    \grad{\mathbb{E}}\sbra{\psi\bra{x_{t}}} = \mathbb{E}\biggl[\grad\psi\bra{x_{t}} + \psi\bra{x_{t}}\grad\biggl( \log K^{\theta}_{0}\bra{k_{0}} - \log p\bra{y_{0:T}} + \\ \sum^{T}_{t=0}\biggl[\log G^{\theta}\bra{y_{t}|x_{t},k_{t}}  + \mathds{1}_{t\geq1}\log K^{\theta}\bra{k_{t}, r_{t-1}}\biggr]\biggr)\biggr] \, ,
\end{multline}
\normalsize
where $\mathds{1}_{t\geq1}$ is defined to be equal to $0$ for $t=0$ and $1$ otherwise.
Since our proposal Eq. \eqref{eq:IMMPFprop} contains no additional parameters, if our gradient estimator is asymptotically equal to \ref{eq:asym-gradient}, then by Theorem \ref{theorem:Consistency} it is consistent. Applying Theorems \ref{theorem:Consistency} and \ref{theorem:weightsnongrad},
\footnotesize
\begin{equation}
\label{eq:grad-of-estimator}
    \grad\sum^{N}_{n=1}\bar{w}^{n}_{T}\psi\bra{x^{n}_{T}} \xrightarrow{L^{2}} \underset{\hat{x}_{T}\sim p\bra{\hat{x}_{T}|y_{0:T}}}{\mathbb{E}}\sbra{\grad\psi\bra{x_{T}} + \psi\bra{x_{t}}\grad{\log\bra{\bar{w}_{T}}}}\, .
\end{equation}
\normalsize
Defining $Z_{T} \defines \sum^{T}_{t=0}\log\sum^{N}_{n=1}w^{n}_{t} - \bra{T+1}\log N$,
\scriptsize
\begin{gather}
\begin{split}
\underset{\hat{x}_{T}\sim p\bra{\hat{x}_{T}|y_{0:T}}}{\mathbb{E}}\sbra{{\psi\bra{x_{T}}\grad\log \bar{w}_{T}}} = \\ - \eE{\psi\bra{x_{T}}}\grad \bra{Z_{T} - Z_{T-1}}  + \underset{\hat{x}_{T}\sim p\bra{\hat{x}_{T}|y_{0:T}}}{\mathbb{E}}\biggl[\psi\bra{x_{T}}\grad\biggl(\log G^{\theta}\bra{y_{T}\mid x_{T},k_{T}} \\ + \frac{1}{p\bra{\hat{x}_{T}\mid y_{0:T-1}}}\underset{\hat{x}_{T-1}\sim p\bra{\hat{x}_{T-1}|y_{0:T-1}}}{\mathbb{E}}\sbra{p\bra{\hat{x}_{T}\mid\hat{x}_{T-1}}\bra{\log \bar{w}_{T-1} + \log K^{\theta}\bra{k_T \mid r_{T-1}}}} \biggr)\biggr]\, 
\end{split}\\
\label{eq:reinforce-grad-of-estimator}
  = -\eE{\psi\bra{x_{T}}}\grad{Z_{T}} + \eE{\psi\bra{x_{T}}\grad\bra{\log K^{\theta}_{0}\bra{k_{0}} +\sum^{T}_{t=0} \log G^{\theta}\bra{y_{t}|x_{t}, k_{t}} + \mathds{1}_{t\geq1}\log K^{\theta}\bra{k_{t}|r_{t-1}}}} \, ,
\end{gather}
\normalsize
where we have derived the final expression by a recursive application of Slutsky's theorem. Note that we cannot marginalise over the past time-steps in the outer expectation since $x_{T}$ is a function of all $x_{0:T-1}$.
\footnotesize
\begin{align}
    \grad{Z_{T}} &\xrightarrow{L^{2}} \eE{\grad\bra{\log K^{\theta}_{0}\bra{k_{0}} +\sum^{T}_{t=0} \log G^{\theta}\bra{y_{t}|x_{t},k_{t}} + \mathds{1}_{t\geq1}\log K^{\theta}\bra{k_{t}|r_{t-1}}}} \\
    \label{eq:ELBO-grad}
    &= \grad \log p\bra{y_{0:T}}, .
\end{align}
\normalsize
Combining \eqref{eq:ELBO-grad}, \eqref{eq:reinforce-grad-of-estimator} and \eqref{eq:grad-of-estimator} we directly obtain \eqref{eq:asym-gradient} so our gradient estimator is consistent. We have proved that estimates of gradients of expectations for bounded functions, by Theorem \ref{theorem:Consistency}, converges in the $L^{2}$ sense to the true expectation under the posterior.
\begin{corollary}
\label{coro:MSE-consist}
    Under the same assumptions as Theorem \ref{theorem:consistency-DIMMPF} the DIMMPF estimate of the gradient of the MSE of a bounded function is weakly consistent. Precisely, for a true state $\tilde{x}_{T}$,
    \footnotesize
    \begin{equation}
        \grad \bra{\sum^{N}_{n=1}\bar{w}^{n}_{T}\psi\bra{x^{n}_{T}} - \psi\bra{\tilde{x}_{T}}}^{2} \xrightarrow{L^{2}} \grad\bra{\mathbb{E}\sbra{\psi\bra{\tilde{x}_{T}}} - \psi\bra{\tilde{x}_{T}}}^{2} \, .
    \end{equation}
    \normalsize
    This result is trivially extended to an average over time-steps and batched trajectories due to the linearity of the gradient operator.
\end{corollary}
\begin{proof}
Define $\tilde{\psi}\bra{x^{n}_{T}} \defines \psi\bra{x^{n}_{T}} - \psi\bra{\tilde{x}_{T}}$.
\footnotesize
\begin{equation}
    \grad \bra{\sum^{N}_{n=1}\bar{w}^{n}_{T}\psi\bra{x^{n}_{T}} - \psi\bra{\tilde{x}_{T}}}^{2} =  \bra{\sum^{N}_{n=1}\bar{w}^{n}_{T} \tilde{\psi}\bra{x^{n}_{t}}}\grad\bra{\sum^{N}_{n=1}\bar{w}^{n}_{T} \tilde{\psi}\bra{x^{n}_{T}}} \, .
\end{equation}
\normalsize
Applying Theorems \ref{theorem:Consistency} and \ref{theorem:consistency-DIMMPF}, as well as Slutsky's theorem we therefore have:
\footnotesize
\begin{multline}
    \grad \bra{\sum^{N}_{n=1}\bar{w}^{n}_{t}\psi\bra{x^{n}_{T}} - \psi\bra{\tilde{x}_{T}}}^{2} \xrightarrow{L^{2}}  \mathbb{E}\sbra{\tilde{\psi}\bra{x^{n}_{T}}}\grad\mathbb{E}\sbra{\tilde{\psi}\bra{x^{n}_{T}}}
    \\
    = \grad\bra{\mathbb{E}\sbra{\psi\bra{x_{T}}} - \psi\bra{\tilde{x}_{T}}}^{2}
\end{multline}
\normalsize
For our case, Eq. \eqref{eq:MSE}, $\psi$ is the identity function, which is not bounded. So, formally we have not proved convergence. However, one could instead take $\psi$ to be an arbitrarily wide rectangular function for a formally consistent estimator.
\footnotesize
\end{proof}
\begin{corollary}
\label{coro:likelihood-consist}
When the weights and their gradient are upper-bounded, the estimate of the gradient of the log-likelihood, is consistent. \Ie
\begin{equation}
    \sum^{T}_{t=0}\grad Z_{T} \xrightarrow{L^{2}} \grad\log p\bra{y_{0:T}} \,.
\end{equation}
\end{corollary}
\begin{proof}
    This result is directly implied by Eq. \eqref{eq:ELBO-grad}. For the data-likelihood, $\sum^{T}_{t=0}\grad Z_{T} = -\grad \mathcal{L}_{\text{data-likelihood}}$, by equation \ref{eq:likelihood-loss}.
\end{proof}

\bibliographystyle{elsarticle-num} 
 \bibliography{ref}
\makeatletter
\def\eq#1{\footnotesize\begin{equation}#1\end{equation}\normalsize}
\def\multi#1{\footnotesize\begin{multline}#1\end{multline}\normalsize}
\def\gather#1{\footnotesize\begin{gather}#1\end{gather}\normalsize}
\makeatother





\end{document}